\documentclass{article}

     \usepackage[final]{amlc_guide}

\usepackage[utf8]{inputenc} 
\usepackage[T1]{fontenc}    
\usepackage{hyperref}       
\usepackage{url}            
\usepackage{booktabs}       
\usepackage{amsfonts}       
\usepackage{nicefrac}       
\usepackage{microtype}      
\usepackage{xcolor}         
\usepackage{hyperref}
\usepackage{url}
\usepackage{booktabs}
\usepackage{amsfonts}
\usepackage{amsmath}
\usepackage{nicefrac}
\usepackage{microtype}
\usepackage{xcolor}
\usepackage{graphicx}
\usepackage{multirow}
\usepackage{subcaption}
\usepackage{framed}
\usepackage{enumitem}
\usepackage{array}
\usepackage{booktabs}
\usepackage{xcolor}
\usepackage{tcolorbox}

\title{Formatting Instructions for AMLC}
\title{\textit{Unlocking Latent Value}: Taxonomy-Guided Recovery of High-Performing Data from Low-Tier Web Corpora}

\author{\textbf{Neeraj Varshney} \quad
\textbf{Sanket Lokegaonkar} \quad \textbf{Nasser Zalmout} \\
\textbf{Qingyu Yin } \quad \textbf{Priyanka Nigam} \quad \textbf{Bing Yin} \\
Amazon \\[2pt]
}

\begin{document}

\maketitle

\begin{abstract}

Dominant web data curation pipelines for pretraining collapse document quality into a single composite score, systematically missing high-value content along dimensions the scorer underweights.
We present a taxonomy-driven framework that recovers this value by filtering along semantically meaningful dimensions that composite scores fail to capture. 
First, building on the ESSENTIAL-WEB taxonomy, we introduce two novel dimensions: \textit{timeliness} and \textit{cultural specificity}, both of which show low pairwise NMI with all existing taxonomy dimensions.
We annotate 14 million web documents using Qwen
2.5 32B and distill these annotations into a lightweight 0.5B model. To enable rapid corpus-wide
annotation, we additionally train a 73M multi-task MLP on E5 embeddings, achieving 50$\times$ the generative model's inference throughput.
Second, to navigate the combinatorial explosion of filter configurations, we introduce a compute-efficient two-pass framework: Pass 1 identifies the strongest individual dimension signals at small scale; Pass 2 constructs and evaluates conjunctive and disjunctive compound filters from the top performers — identifying high-performing configurations at a fraction of the cost of full scaling-law experiments. 
Applying the selected filters to deprioritized web data, taxonomy-filtered subsets consistently outperform their unfiltered baselines and surpass even the highest-quality tier on most benchmarks. On mid-tier data, our best filter improves over its unfiltered baseline by an average of 12.1\% on reasoning, 9.5\% on coding, and 2.0\% on knowledge benchmarks, exceeding unfiltered top-tier data by 6.7\% on reasoning and 13.7\% on coding. 
Furthermore, filtered data from two tiers below the typical production threshold improves by 22.3\% on reasoning and 19.5\% on coding over its unfiltered baseline and even surpassing the top-tier data on coding benchmarks.
These results establish that vast latent value remains locked in deprioritized web data, and that multi-dimensional taxonomy filtering is a principled, compute-efficient key to unlocking it.

\end{abstract}

\section{Introduction}
\label{sec:intro}

The value of a web document for LLM pretraining is not a scalar property — it is multi-dimensional, and the right filter depends on the target capability. Yet dominant pretraining pipelines treat it as exactly that: each document receives a single composite quality score, and training mixes draw disproportionately from the highest-ranked tiers while discarding or deprioritizing the rest \citep{dclm2024, fineweb2024}.

The problem with this approach is fundamental. A composite score — whether from a single classifier or an ensemble — conflates distinct document properties such as educational depth, reasoning complexity, temporal relevance, and cultural scope into an opaque scalar. This prevents curators from asking targeted questions such as ``retain only documents with deep reasoning that carry evergreen content''. Documents that fall below the selected threshold are either discarded or rewritten without any understanding of why they scored low or which of their properties make them valuable. Rewriting in particular applies expensive LLM compute uniformly across the tier \citep{nemotroncc2024}, without distinguishing documents that carry genuine latent value from those that do not.

ESSENTIAL-WEB \citep{essentialweb2025} (EAI) addressed the opacity problem by introducing a twelve-field taxonomy spanning cognitive complexity, format, topic, and data quality artifacts, and demonstrated that filtering along individual dimensions produces competitive domain-specific subsets. 
However, EAI focused exclusively on single-dimension filters and manual compound combinations. It did not systematically explore the combinatorial space of multi-dimensional filter configurations 
nor did it investigate whether taxonomy-driven curation can recover value from data that quality ranking deprioritizes.

In this work, we address both gaps. Our contributions are threefold:

\textbf{Extended taxonomy and efficient annotation.} We augment the twelve EAI dimensions with two novel dimensions: \textit{timeliness} (the temporal durability of a document's informational value) and \textit{cultural specificity} (the degree to which content is tied to specific cultural or geographic contexts). 
Both dimensions show low pairwise NMI with all existing taxonomy dimensions (timeliness: max NMI $= 0.21$, cultural specificity: max NMI $= 0.11$), confirming they capture genuinely novel signal not already encoded in existing taxonomy.
We annotate 14 million web documents using Qwen 2.5 32B and distill these annotations into a lightweight 0.5B model.
To enable rapid corpus-wide annotation, we additionally train a 73M multi-task MLP on E5 embeddings \citep{e5multilingual}, achieving high exact-match accuracy with the generative model at 50$\times$ its throughput. 
Our comprehensive experiments demonstrate that timeliness emerges as the most impactful dimension.

\textbf{Compute-efficient two-pass experimental framework.} 
Searching the space of compound filters over a large number of ordinal dimensions with multiple threshold values per dimension is combinatorially explosive. We introduce a two-pass framework that both discovers promising compound filter combinations and screens them efficiently. Pass 1 evaluates each dimension independently at small scale to identify the strongest individual signals; Pass 2 constructs conjunctive and disjunctive compound filters from the top-performing dimensions and evaluates them at the same scale. This systematic construction process surfaces non-obvious combinations that manual search would miss, at a fraction of the cost of full scaling-law experiments.
The consistent ordering between our screening and full scaling-law results validates the methodology.

\textbf{Value recovery from deprioritized data.}
Applying our framework to web data that conventional pipelines deprioritize,
taxonomy-filtered subsets consistently outperform the unfiltered baselines across reasoning, knowledge, and coding benchmarks. On mid-tier data, our overall best compound filter improves over its unfiltered baseline by an average of 12.1\% on reasoning and 9.5\% on coding, and exceeds unfiltered top-tier data by 6.7\% on reasoning and 13.7\% on coding. The effect extends further down the quality spectrum: filtered data from two tiers below the typical production threshold (data that standard pipelines often discard entirely) improves over its unfiltered baseline by an average of 22.3\% on reasoning and 19.5\% on coding, while surpassing unfiltered top-tier data on coding benchmarks. Our analysis reveals a systematic tradeoff between filter stringency and capability profile: strict filters favor numerical reasoning and coding, while broader filters better preserve factual knowledge breadth, directly motivating principled dataset construction for staged training.

\section{Background and Related Work}
\label{sec:background}
Our work sits at the intersection of four research areas: web data quality filtering, scaling-law experimentation for data curation, document taxonomies, and the temporal and cultural properties of training data. We review each in turn and position our contributions relative to prior work.

\textbf{Web Data Quality Filtering.} Modern LLM pretraining pipelines rely heavily on quality filtering to select high-value subsets from large web crawls. DCLM \citep{dclm2024} showed that a fastText classifier trained on high-quality reference corpora can effectively score documents via simple threshold filtering. FineWeb \citep{fineweb2024} and FineWeb-Edu \citep{finewebedu2024} refined this with educational quality classifiers, showing that domain-specific quality signals outperform general-purpose ones for targeted capabilities. NemotronCC \citep{nemotroncc2024} ensembled three classifiers into a composite quality score and further applied LLM-based rewriting to recover value from lower-scored documents. More recent work has explored GPT-4o-based filtering \citep{biderman2023pythia} as a high-precision but expensive alternative.
Despite their differences, all reduce document quality to a single composite score, leaving substantial latent value inaccessible — the value our framework is designed to recover.

\textbf{Scaling-Law Experiments for Data Curation.} The standard approach to evaluating data mixtures uses scaling-law ladders \citep{chinchilla}: training at multiple compute scales and extrapolating to a target budget. DCLM applied this to compare data curation strategies, but the approach becomes prohibitively expensive when the candidate space is large — as it is when exploring compound filters over many ordinal dimensions. Prior work has largely addressed this by restricting the search space manually or evaluating only a small number of pre-selected configurations. 
Our two-pass framework takes a different approach: rather than manually restricting the search space, it uses small-scale runs as a principled screening stage to systematically rank individual dimensions, construct promising compound combinations from the top performers, and promote only the most promising configurations.

\textbf{Multi-Dimensional Web Taxonomies.} ESSENTIAL-WEB \citep{essentialweb2025} introduced the first large-scale multi-dimensional taxonomy for web data curation, with twelve fields spanning cognitive complexity, document format, topic classification, and data quality artifacts. It demonstrated that filtering along individual dimensions produces competitive domain-specific subsets and distilled a 32B teacher into a lightweight 0.5B classifier. 
However, EAI focused exclusively on single-dimension filters, leaving systematic exploration of compound multi-dimensional configurations and value recovery from deprioritized data unexplored.
Our work directly addresses both gaps: we extend the taxonomy with two novel dimensions, introduce a systematic framework for compound filter discovery, and demonstrate value recovery from deprioritized tiers.

\textbf{Temporal and Cultural Properties of Training Data.} To our knowledge, timeliness and cultural specificity have not previously been introduced as explicit curation dimensions for LLM pretraining. Prior work on temporal generalization \citep{10.5555/3540261.3542508} studied how models degrade on out-of-distribution time periods, but from an evaluation perspective rather than a data curation one.
Our work reframes timeliness as a curation dimension, providing the first empirical demonstration that filtering on temporal durability has a measurable and substantial impact on downstream pretraining performance.
On the cultural dimension, \cite{lucy2024aboutme} showed that standard quality filters carry implicit cultural biases. This motivates explicit modeling of cultural scope as a curation dimension.

\section{Extended Taxonomy}
\label{sec:taxonomy}

We extend the EAI twelve-dimension taxonomy (Appendix \ref{app_all_dimensions_section}) with two novel general-purpose dimensions: \textit{timeliness} and \textit{cultural specificity}. These dimensions were identified as meaningful gaps and properties that are intuitively predictive of downstream training value. 
Both are defined as six-level ordinal scales (Table~\ref{tab:new_dimensions_detailed}) and participate directly in the compound filtering experiments. The existing categorical dimensions of EAI — FDC topic hierarchy and document type — serve as analytical dimensions for content characterization and cross-analysis rather than filtering targets.

\subsection{Timeliness}

Timeliness captures the temporal durability of a document's informational value. We define a six-level ordinal scale ranging from \textit{highly time-sensitive} (breaking news, stock prices, real-time updates) to \textit{completely evergreen} (scientific laws, mathematical concepts, how-to guides), with an additional \textit{indeterminate} category for documents that cannot be classified.
Timeliness proves to be the most impactful new dimension: it appears as a conjunctive clause in nearly all high-performing compound filters discovered in our work (\S\ref{sec:pass2_results}), in fact, just timeliness-filtered mid-tier data outperforms even unfiltered top-tier data on most benchmarks (\S\ref{sec:new_dimensions}).

The distribution of document types across timeliness values validates the dimension's semantics: highly time-sensitive content is dominated by News/Editorial documents (88.9\%), while completely evergreen content is dominated by Reference/Educational documents (84.3\%), with Knowledge Articles, Tutorials, and Q\&A Forums rising monotonically across the scale. 
Critically, completely evergreen content contains significantty more Intermediate+Advanced reasoning content compared to highly time-sensitive content (32.4\% vs. 6.3\%), providing a mechanistic explanation for why timeliness filtering improves downstream performance on reasoning benchmarks.
We study the distribution analysis in detail in Appendix \ref{appendix_sec_cross_analysis} and Tables \ref{tab:timeliness_doctype} - \ref{tab:timeliness_edu}.

\subsection{Cultural Specificity}

Cultural specificity captures the degree to which a document's content is tied to specific cultural norms or geographic contexts. We define a six-level scale ranging from \textit{highly localized} (regional news, local government notices, culturally-bound practices) to \textit{completely universal} (mathematical concepts, formal scientific principles, abstract theoretical frameworks).

Completely universal content concentrates strongly in Reference/Educational (72.1\%) and Academic/Research (17.2\%) documents, while highly localized content spans a broader mix of genres including News/Editorial (26.\%) — confirming that universal content is predominantly formal reference and scientific material (Appendix \ref{appendix_sec_cross_analysis}, Table \ref{tab:cultural_doctype}).

Strict universal filtering reduces knowledge breadth on benchmarks such as MMLU that test culturally-contextualized domains including law, social sciences, and regional history — a tradeoff we analyze in \S\ref{sec:new_dimensions}.
Cultural specificity is also particularly valuable for non-English data curation, where it can prioritize documents containing genuinely local knowledge not already covered by English-language sources, making it a complementary signal to timeliness.

\subsection{Annotation Pipeline and Embedding-Based Multi-Task Classifier}

We annotate 14 million diverse web documents using Qwen 2.5 32B with chain-of-thought prompts (\ref{app_timeliness_prompt} and \ref{app_cultural_prompt}) for the new dimensions and the original 0.5B model directly for the existing EAI dimensions.
The 0.5B distilled model is then further fine-tuned on our new annotations following the original distillation pipeline, producing an updated 0.5B model covering all 14 dimensions (Table \ref{tab:taxonomy_dimensions}).

Generative classifiers are accurate but computationally expensive at web scale. To enable rapid experimentation across tens of billions of documents, we train a 73M-parameter multi-task MLP on pre-computed E5-multilingual-large-instruct embeddings \citep{e5multilingual} (1024-dim, 512-token context). Because embeddings are computed once and reused, new filter configurations can be scored and evaluated within hours rather than days.

The aggregate primary exact-match accuracy gap is just -7.5\%. Crucially, dimensions predominantly detectable from opening paragraphs like timeliness (-2.1\%) and cultural specificity (-2.6\%) show minimal degradation, while full-document dimensions like document type V2 (-17.4\%) show relatively large gaps — confirming the embedding classifier is well-suited for the dimensions most central to our filtering framework. Full architecture and training details are provided in Appendix \ref{app:genvdisc}.

\section{Two-Pass Experimental Framework}
\label{sec_two_pass}

Our framework addresses the combinatorial explosion of compound filter configurations through a two-stage process: Pass 1 identifies the strongest individual dimension signals at low cost, and Pass 2 uses those signals to construct and evaluate promising compound combinations — systematically exploring a space that would be intractable to search exhaustively.
We adopt scaling-law notation where $L_n$ denotes training run at the $n$-th rung of the compute ladder. 
$L_{10}$ (trained on 42.8B tokens) serves as the screening scale, with performance measured via \textit{answer loss} (lower is better).

\textbf{Pass 1: Independent Dimension Evaluation.}
Pass 1 trains models on subsets filtered to each dimension value independently
at $L_{10}$ scale with two random seeds, measuring answer loss across the full benchmark suite.
(Dimension-value) tuples are ranked by their average improvement over the unfiltered baseline, producing a ranked list of the most informative curation signals. This ranking serves two purposes: it identifies which dimensions are worth including in compound filters, and it reveals which dimension values are actively harmful — providing negative filters that compound configurations should exclude. The top-performing dimensions are carried forward to Pass 2.

\textbf{Pass 2: Compound Filter Evaluation.}
Pass 1 produces a ranked list of the strongest individual signals; Pass 2 asks whether combining them yields further gains and, if so, which combinations are most effective. The 12 compound filters (Table~\ref{tab:filters_retentions}) are constructed from the top-5 Pass 1 dimensions — educational level, Bloom cognitive, Bloom knowledge, reasoning depth, and timeliness — varying two principled axes: (a) combination logic, ranging from pure disjunctive (broad coverage, high retention) to pure conjunctive (high precision, low retention), to understand the selectivity–performance tradeoff; and (b) whether each combination is further intersected with timeliness, motivated by its status as the single strongest Pass 1 signal, to test whether temporal durability adds orthogonal value on top of content-quality dimensions. This construction enables controlled comparisons. $L_{10}$ evaluations are run for all 12 configurations before selecting filters for full scaling-law validation.

\textbf{Compute Efficiency}
The two-pass framework makes systematic exploration of the compound filter space tractable. A naive approach — running a full $L_4$-through-$L_{12}$ scaling-law ladder for each possible configuration — would require thousands of training runs. By contrast, both passes operate exclusively at $L_{10}$ scale, with full ladder runs reserved for post-hoc validation of the selected filter. This design reduces the total compute required for systematic compound filter search significantly, while the consistent ordering between $L_{10}$ screening and full-ladder results confirms that $L_{10}$ is a reliable proxy for larger-scale performance.

\section{Experiments and Results}
\label{sec:experiments}

\subsection{Experimental Setup}

\textbf{Data source.} Our experiments are conducted on a large-scale proprietary web corpus partitioned into 200 quality buckets using a 3-model ensemble quality classifier as NemotronCC, where bucket 1 is lowest quality and bucket 200 is the highest. We focus on two tiers that conventional pipelines deprioritize. First, buckets 191–199 — the second-highest quality tier, above the corpus median but below bucket 200, which is used primarily in conventional production training mixes. This tier represents data that is close to production quality but systematically excluded by threshold-based pipelines. Second, buckets 150–190, comprising $\sim5x$ the tokens of buckets 191-199 — data that sits further below the typical production threshold and that standard pipelines often discard entirely. Evaluating both tiers allows us to assess both the standalone impact of taxonomy filtering and the extent to which it recovers value across the quality spectrum.

\textbf{Benchmarks.} We evaluate on reasoning, coding, and knowledge benchmarks: GSM8K~\citep{gsm8k}, Minerva-Math~\citep{minerva}, MMLU~\citep{mmlu}, HumanEval~\citep{humaneval}, and MBPP~\citep{mbpp}. Performance is reported as answer loss for scaling-law figures, providing a continuous signal that is more sensitive to data quality differences than discrete accuracy metrics at small compute scales.

\textbf{Compute ladder.} Rungs span $L_4$ (13.8B tokens) through $L_{12}$ (56.6B tokens), with $L_{10}$ (42.8B tokens) serving as the screening scale for Pass 1 and Pass 2 evaluations. Full ladder runs span $L_4$–$L_{12}$ and are used for post-hoc validation of selected filter configurations.

\subsection{Timeliness and Cultural Specificity as Standalone Filtering Signals}
\label{sec:new_dimensions}

We evaluate timeliness and cultural specificity as standalone filtering dimensions, each applied independently to various bucket tiers and compared against the unfiltered baseline. Since these dimensions are absent from conventional quality scoring pipelines, any improvement they yield represents genuinely new signal that scalar quality scores do not provide.

\begin{figure}[t]
\centering
\begin{subfigure}[b]{\linewidth}
\includegraphics[width=\linewidth]{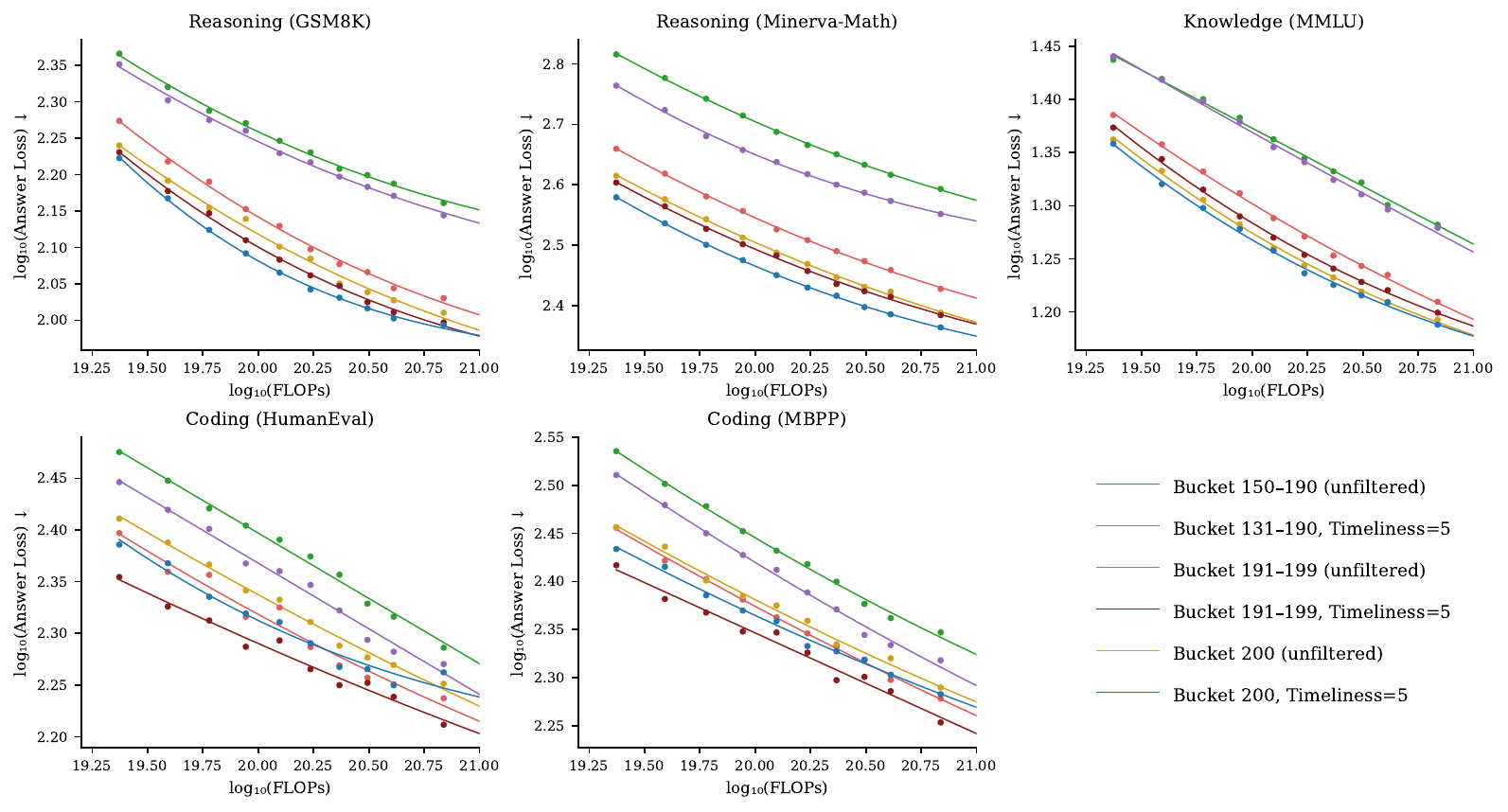}
\end{subfigure}
\caption{Scaling-law curves comparing unfiltered data against subsets filtered to timeliness = 5 (completely evergreen) across the three quality tiers. Within each tier, timeliness filtering consistently improves over the unfiltered baseline across reasoning, math, and coding benchmarks.}
\label{fig:timeliness}
\end{figure}

\paragraph{Timeliness.} 
Figure \ref{fig:timeliness} shows the impact of timeliness filtering on the three bucket tiers.
Within each tier, timeliness filtering consistently improves over the unfiltered baseline — including $B_{200}T_5$ over $B_{200}$, confirming that even the highest-quality bucket contains time-sensitive content that degrades training signal.

Crucially, $B_{191\text{--}199}T_5$ outperforms unfiltered top-tier data ($B_{200}$) on coding and reasoning benchmarks, establishing that a single taxonomy dimension can recover value that quality bucketing misses entirely. This is the central finding of this subsection: data that conventional pipelines deprioritize in favor of bucket 200 exceeds it simply by filtering for temporal durability.

The value-recovery effect extends further down the quality spectrum. $B_{131\text{--}190}T_5$ not only outperforms $B_{131\text{--}190}$ on all benchmarks, but approaches the performance of $B_{191\text{--}199}$ on some benchmarks — confirming that the timeliness filter, not the source data tier, drives the improvement. On MMLU, timeliness filtering shows relatively smaller impact on the lower tier, likely because broad factual knowledge is distributed across both evergreen and time-sensitive content at that quality level.

\paragraph{Cultural Specificity.} 
Figure \ref{fig:cultural} shows the impact on cultural specificity filtering on buckets 191–199.
Filtering for completely universal content consistently improves performance across reasoning and coding benchmarks — GSM8K, Minerva-Math, HumanEval, and MBPP all show clear gains over the unfiltered baseline. The one exception is MMLU, where the unfiltered baseline shows a marginal increase in loss. This regression is attributable to knowledge breadth reduction: MMLU tests diverse domains including culturally-contextualized topics such as law, social sciences, and regional history, which strict universal filtering removes alongside genuinely low-value locally-specific content.

Cultural specificity thus presents a more nuanced tradeoff than timeliness: it is a reliable signal for reasoning and coding capability but comes at a small cost to broad factual knowledge coverage. For practitioners targeting reasoning-heavy training stages, cultural specificity $= 5$ filtering is a straightforward win. For stages requiring broad knowledge retention, a mixture-based approach that preserves some locally-specific content may be preferable.

\subsection{Pass 1 Results - Identifying Strong Individual Filtering Configurations} 
\label{sec:pass1_results}

\begin{figure}[t]
\centering
\begin{subfigure}[b]{0.9\linewidth}
\includegraphics[width=\linewidth]{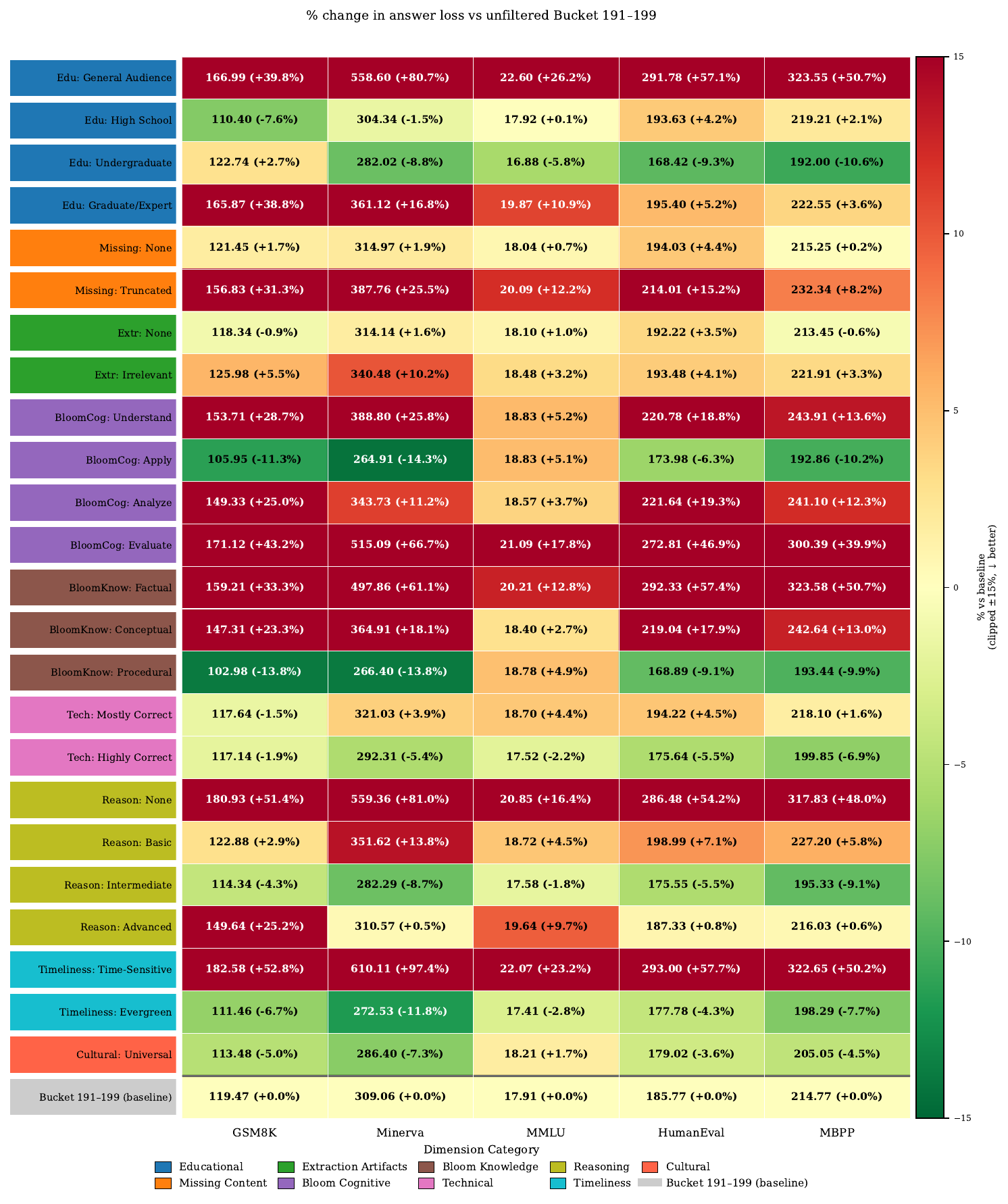}
\end{subfigure}
\caption{Pass 1 Individual dimension value results for Bucket 191-199: Percentage change in answer loss relative to the unfiltered bucket 191–199 baseline for each individual taxonomy dimension value. }
\label{fig:pass_1}
\end{figure}

Figure~\ref{fig:pass_1} shows percentage change in answer loss relative to the unfiltered bucket 191–199 baseline for each individual taxonomy dimension value.

\textbf{Top-performing dimensions.} Educational level $\geq 2$, Timeliness $= 5$ (Completely Evergreen), Bloom cognitive $=$ Apply, Bloom knowledge $=$ Procedural, technical correctness $=$ Highly Correct, Cultural specificity $=$ Universal, and reasoning depth generally outperform the baseline across all reasoning, math, and coding benchmarks. On MMLU, absolute differences between dimension values are smaller, but the same dimensions maintain their relative advantage.
This suggests a consistent underlying signal: documents that require active application of procedural knowledge are systematically more valuable than purely factual or deeply analytical content. This aligns with the timeliness finding — evergreen how-to guides and scientific explanations tend to be both procedural and temporally stable.
Bloom cognitive $= 3$ (Apply) and Bloom knowledge $= 3$ (Procedural) are strong on reasoning and math benchmarks specifically. Technical correctness $= 4$ (Highly Correct) shows particular strength on coding benchmarks (HumanEval, MBPP), suggesting that syntactic and semantic correctness is especially valuable for code generation capability.

\textbf{Harmful dimension values.} Several dimension values substantially hurt performance. Timeliness $= 2$ (Predominantly Time-Sensitive) is the most damaging on Minerva-Math. This is the starkest finding in Pass 1: time-sensitive content is not merely unhelpful but actively harmful for mathematical reasoning training. Reasoning depth $= 1$ (No Reasoning) similarly produces very high loss on Minerva-Math and coding benchmarks. Educational level $= 1$ (General Audience) is the worst-performing educational level across all benchmarks. Bloom cognitive $=$ Analyze and $=$ Evaluate, and Bloom knowledge $=$ Conceptual also consistently underperform, suggesting that purely theoretical content without procedural application is less valuable than content requiring active knowledge use.
These results confirm that harmful dimension values actively introduce low-quality training signal that degrades model capability rather than being merely neutral.

\textbf{Data quality artifacts.} Missing content $=$ Truncated Snippets hurts performance across benchmarks, confirming that documents with incomplete extraction are a meaningful source of noise. Extraction artifacts show a similar but weaker pattern. 

\textbf{Dimensions carried forward to Pass 2.} Based on Pass 1 results, we carry forward five dimensions for compound filter construction: educational level $\geq 2$, Bloom cognitive $=$ Apply, Bloom knowledge $=$ Procedural, timeliness $= 5$, and reasoning depth $\geq 3$. These dimensions show the strongest and most consistent improvements across the benchmark suite.

\subsection{Pass 2 Results - Identifying Strong Compound Filtering Configurations} 
\label{sec:pass2_results}

\begin{figure}[t]
\centering
\begin{subfigure}[b]{0.95\linewidth}
\includegraphics[width=\linewidth]{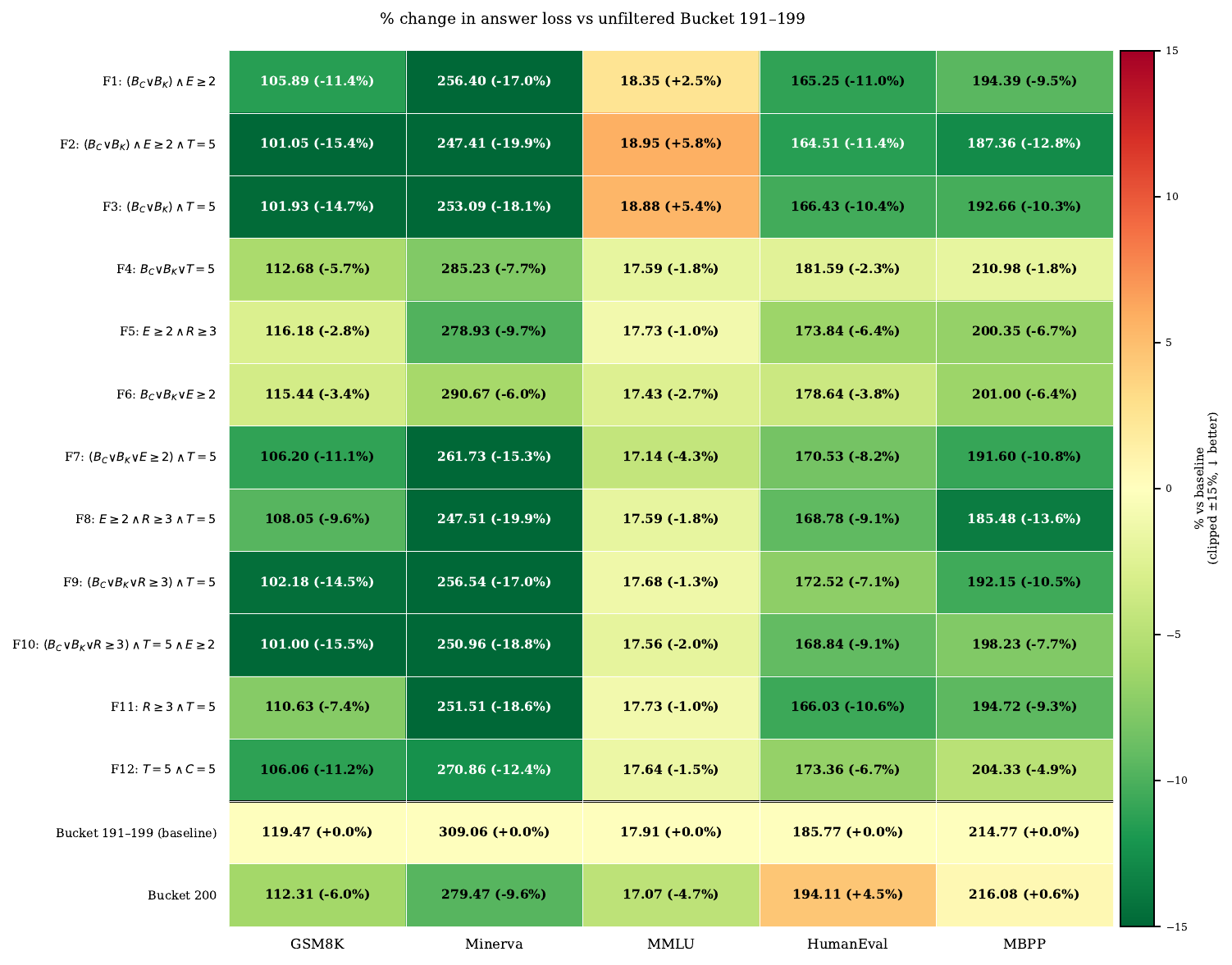}
\end{subfigure}
\caption{Pass 2 Compound filter results for Bucket 191-199: Percentage change in answer loss relative to the unfiltered bucket 191–199 baseline for compound conjunctive and disjunctive filter configurations.}
\label{fig:pass_2}
\end{figure}

Figure~\ref{fig:pass_2} presents the percentage change in answer loss relative to the unfiltered $B_{191\text{--}199}$ baseline for all 12 compound filter configurations (F1 -- F12, Table~\ref{tab:filters_retentions}) as described in \S\ref{sec_two_pass}.

\textbf{Compound filters beat both baselines on reasoning and coding.} The improvement over the $B_{191\text{--}199}$ baseline is consistent and substantial across all filter configurations. Critically, almost all compound filters also outperform unfiltered $B_{200}$ on GSM8K, Minerva-Math, HumanEval, and MBPP — extending the finding from §5.2 that taxonomy-filtered mid-tier data surpasses conventional top-tier production data across the full range of compound configurations tested.

\textbf{F8 is the strongest overall filter.} 
F8 ($E{\geq}2 \wedge R{\geq}3 \wedge T{=}5$) with $10.3\%$ retention achieves the best average improvement across all five benchmarks (at $-10.8\%$) and is the top performer on both Minerva-Math ($-19.9\%$) and MBPP ($-13.6\%$). 
It combines three complementary signals — educational level, reasoning depth, and timeliness — each of which Pass 1 identified as independently strong, and their conjunction concentrates the highest-value documents in the tier.

\textbf{Timeliness as a conjunctive clause strengthens reasoning and math.} 
Comparing matched filter pairs that differ only in the presence of the $T{=}5$ constraint directly isolates its contribution. Adding $T{=}5$ to $E{\geq}2 \wedge R{\geq}3$ (F5 → F8) improves Minerva-Math by 31.4 points and GSM8K by 8.1 points, while reducing retention from 33.5\% to 10.3\%. This establishes that timeliness is not a proxy for educational or reasoning depth — it captures a distinct document property that the other dimensions do not. A document can be highly educational and require intermediate reasoning while still being time-sensitive (e.g., a detailed analysis of a recent policy change), and filtering it out improves training signal on reasoning benchmarks. On MMLU, adding $T{=}5$ has a marginal improvement (-0.8\%), consistent with the knowledge-breadth tradeoff discussed later in this subsection.

\textbf{Selectivity and capability profile.} Sorting filters by data retention rate reveals a broadly monotonic relationship on reasoning benchmarks — stricter filters consistently achieve lower Minerva-Math and GSM8K losses — but composition matters alongside selectivity. At similar retention levels, filters that combine timeliness with educational and reasoning constraints outperform those that do not: F7 ($23.3\%$ retention, Minerva loss $261.7$) outperforms F12 ($22.6\%$, $270.9$) despite nearly identical data volume, and F10 ($14.3\%$, $251.0$) marginally beats both F11 ($10.6\%$, $251.5$) and F3 ($10.8\%$, $253.1$) despite retaining more data. 
On MMLU, the pattern inverts: broader filters consistently outperform strict ones, directly reflecting MMLU's nature as a breadth benchmark where factual knowledge is distributed across document types that strict filters systematically exclude.
Notably, filters that include timeliness are neutral to beneficial on MMLU — F7 improves over baseline by 4.3\% and F8 by 1.8\% — while strict Bloom+educational constraints (F2, F3) show the largest regressions (+5.8\%, +5.4\%). This tradeoff directly motivates staged training: broader filters for knowledge-building phases, stricter conjunctive filters for reasoning-focused annealing. Based on these L10 screening results, we select F8 for full scaling-law validation in \S\ref{sec:full_scaling_law_ladder}.

\subsection{Full Scaling Law Ladder Runs}
\label{sec:full_scaling_law_ladder}

\begin{figure}[t]
\centering
\begin{subfigure}[b]{\linewidth}
\includegraphics[width=\linewidth]{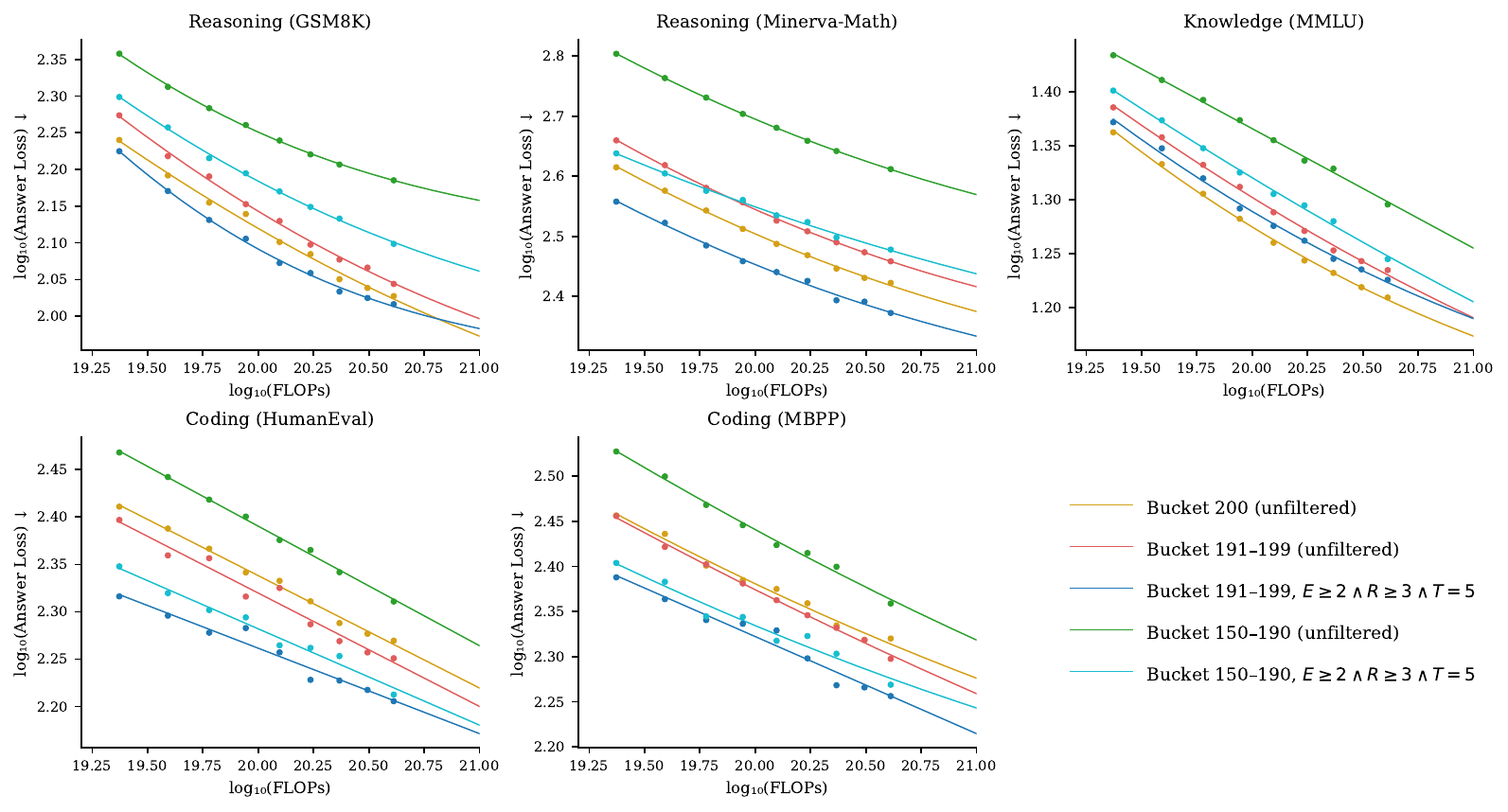}
\end{subfigure}
\caption{Full scaling-law ladder results for F8 ($E{\geq}2 \wedge R{\geq}3 \wedge T{=}5$) applied to mid-tier (buckets 191–199) and lower-tier (buckets 150–190) data compared against various unfiltered baselines.}
\label{fig:full_scaling_law_ladder_results}
\end{figure}

The $L_{10}$ screening results identified F8 ($E{\geq}2 \wedge R{\geq}3 \wedge T{=}5$, 10.3\% retention) as the strongest overall compound filter. We validate it using full scaling-law ladders spanning $L_4$ through $L_{12}$, alongside unfiltered bucket 191–199 and bucket 200 baselines (Figure~\ref{fig:full_scaling_law_ladder_results}). The ordering observed at $L_{10}$ is preserved across the full compute range, confirming that $L_{10}$ screening is a reliable proxy for larger-scale performance.

\textbf{Reasoning and coding gains are substantial and consistent.} At $L_{12}$, F8 achieves a Minerva-Math loss of 235.7 — an 18.0\% improvement over unfiltered bucket 191–199 and 10.9\% over bucket 200 — and a GSM8K loss of 103.9 (-6.2\% and -2.5\% respectively). The coding gains are even stronger: F8 achieves HumanEval loss of 160.5 and MBPP loss of 180.4, surpassing unfiltered bucket 200 by 13.7\% on both benchmarks. This is the central result: filtered mid-tier data consistently outperforms the conventional top-tier across reasoning and coding at every compute scale. On MMLU, F8 improves over the bucket 191–199 baseline (16.82 vs. 17.17, -2.0\%) while falling slightly short of bucket 200 (16.20), consistent with the knowledge-breadth tradeoff identified in §5.4.

\textbf{Value recovery extends further down the quality spectrum.} Applying F8 to buckets 150–190 — data that standard pipelines often discard entirely — reduces GSM8K loss from 153.2 to 125.5 (-18.1\%) and Minerva-Math loss from 409.1 to 300.6 (-26.5\%). On coding, filtered bucket 150–190 data achieves HumanEval loss of 163.1 and MBPP loss of 185.7, surpassing unfiltered bucket 200 by 12.3\% and 11.1\% respectively. The fact that data two tiers below the production threshold can surpass the top tier after filtering establishes that the bottleneck in web data curation is not the quantity of high-quality data but the precision of the lens used to find it.

\section{Conclusion}
\label{sec:conclusion}

In this work, we have shown that the value of a web document for LLM pretraining is not a scalar property — it is multi-dimensional, and the right filter depends on what capability you are trying to build. Our taxonomy-driven framework makes this actionable: by extending the ESSENTIAL-WEB taxonomy with two novel dimensions, introducing a compute-efficient two-pass framework for compound filter discovery, and applying the resulting filters to deprioritized web data, we recover substantial training value that scalar quality pipelines systematically miss.
A single new dimension — timeliness — is sufficient to make mid-tier data surpass unfiltered top-tier data on reasoning and coding; compound filters push this further, with gains of up to 10.9\% on Minerva-Math and 13.7\% on coding over the top tier, and the effect extends two tiers below the production threshold with 22.3\% and 19.5\% improvements on reasoning and coding respectively. A second finding is the systematic tradeoff between filter stringency and capability profile: strict conjunctive filters excel on reasoning and coding while broader disjunctive filters better preserve factual knowledge breadth, directly motivating staged training with broader filters for knowledge-building phases and stricter filters for reasoning-focused annealing.

\textbf{Limitations.} Our work is primarily validated on English-language web data; the utility of the new dimensions — particularly cultural specificity — for multilingual curation remains to be demonstrated at scale. 

\bibliographystyle{plain}
\bibliography{references}


\newpage
\appendix

\section*{Appendix}

\section{Taxonomy Dimensions}
\label{app_all_dimensions_section}

Table \ref{tab:taxonomy_dimensions} shows the essential web taxonomy dimensions and Table \ref{tab:new_dimensions_detailed} shows the scale definitions for the two novel taxonomy dimensions introduced in this work.

\begin{table}[h]
\centering
\small
\caption{ESSENTIAL-WEB taxonomy dimensions and their descriptions, extended with two 
novel dimensions introduced in this work (marked with $\dagger$).}
\label{tab:taxonomy_dimensions}
\begin{tabular}{p{0.38\linewidth} p{0.54\linewidth}}
\toprule
\textbf{Description} & \textbf{Categories} \\
\midrule

\textbf{FDC.} Free Decimal Correspondence is a 
public-domain analogue of the Dewey Decimal System. 
Labels subject matter hierarchically. &
\textbf{Level 1}: broad topic (ex: \texttt{5 - Science}) \newline
\textbf{Level 2}: fine topic (ex: \texttt{51 - Mathematics}) \newline
\textbf{Level 3}: very fine topic (ex: \texttt{512 - Algebra}) \\

\midrule

\textbf{Bloom.} Educational-objective taxonomy. &
\textbf{Cognitive Process}: mental effort required \newline
\textbf{Knowledge Domain}: content abstraction \\

\midrule

\textbf{Document Type.} Web page types. &
\textbf{V1}: broad types (17 categories) \newline
\textbf{V2}: fine types (25 categories) \\

\midrule

\textbf{Content Quality.} Measures rigor and target audience. &
\textbf{Reasoning Depth}: thinking steps required \newline
\textbf{Educational Level}: reader background assumed \newline
\textbf{Technical Correctness}: factual and syntactic correctness \\

\midrule

\textbf{Extraction.} Crawl artifacts and text extraction errors. &
\textbf{Extraction Artifacts}: HTML extraction errors \newline
\textbf{Missing Content}: from scrape or extraction \\

\midrule

\textbf{Timeliness}$^\dagger$\textbf{.} Temporal durability of a 
document's informational value. &
6-level ordinal scale: \textit{Highly Time-Sensitive} (1) $\to$ 
\textit{Completely Evergreen} (5) $+$ Indeterminate (6) \\

\midrule

\textbf{Cultural Specificity}$^\dagger$\textbf{.} Degree to which 
content is tied to specific cultural or geographic contexts. &
6-level ordinal scale: \textit{Highly Localized} (1) $\to$ 
\textit{Completely Universal} (5) $+$ Indeterminate (6) \\

\bottomrule
\multicolumn{2}{l}{$^\dagger$ Novel dimensions introduced in this work.}
\end{tabular}
\end{table}

\begin{table}[t]
\caption{Scale definitions for the two novel taxonomy dimensions introduced in this work.}
\label{tab:new_dimensions_detailed}
\centering
\footnotesize
\setlength{\tabcolsep}{3pt}
\renewcommand{\arraystretch}{0.95}

\begin{tabular}{@{}p{3.1cm}p{3.4cm}p{3.1cm}p{3.4cm}@{}}
\toprule
\multicolumn{2}{c}{\textbf{Timeliness}} & \multicolumn{2}{c}{\textbf{Cultural Specificity}} \\
\cmidrule(lr){1-2} \cmidrule(lr){3-4}
Label & Definition & Label & Definition \\
\midrule
Highly Time-Sensitive &
Breaking news, stock prices, real-time updates; outdated within hours or days &
Highly Localized &
Regional news, local reviews, cultural festivals; context-dependent \\

Pred.\ Time-Sensitive &
Quarterly reports, sports seasons, product launches; outdated within weeks or months &
Mostly Local/Regional &
National laws, country policies, regional business practices \\

Balanced &
Event-specific content with enduring principles &
Balanced Local-Global &
Global issues illustrated via local examples \\

Predominantly Evergreen &
General knowledge with minor time-bound elements &
Global but Contextualized &
Globally relevant but reflects cultural assumptions/examples \\

Completely Evergreen &
Scientific laws, math concepts, history, how-to guides &
Completely Universal &
Math, science, abstract frameworks; culture-neutral \\

Indeterminate &
Cannot be classified &
Indeterminate &
Cannot be classified \\
\bottomrule
\end{tabular}
\end{table}

\section{Timeliness Dimension}

\subsection{LLM Annotation Prompt}
\label{app_timeliness_prompt}

\begin{tcolorbox}[title=Timeliness Annotation Prompt]
\small

\textbf{\#\# [Custom] Timeliness Taxonomy}

\textbf{\#\#\# Taxonomy Definition}

\textbf{1 Highly Time-Sensitive}
\begin{itemize}[noitemsep]
\item Content focuses exclusively on current events, breaking news, or real-time updates.
\item Examples: stock prices, weather reports, breaking news, live event coverage.
\item Information would likely be outdated within hours or days.
\end{itemize}

\textbf{2 Predominantly Time-Sensitive}
\begin{itemize}[noitemsep]
\item Content primarily focused on recent events or developments.
\item Examples: quarterly business reports, sports season coverage, current trends, product launches.
\item Would become outdated within weeks or months.
\end{itemize}

\textbf{3 Balanced Time-Sensitive and Evergreen}
\begin{itemize}[noitemsep]
\item Content equally balances timely and evergreen elements.
\item References specific events or dates but includes enduring principles or concepts.
\item Some aspects remain relevant beyond the immediate timeframe.
\end{itemize}

\textbf{4 Predominantly Evergreen}
\begin{itemize}[noitemsep]
\item Content contains mostly general knowledge, fundamental concepts, or established facts.
\item May include occasional references to dates or events, but these aren't central.
\item Long-term relevance with minor time-bound elements.
\end{itemize}

\textbf{5 Completely Evergreen}
\begin{itemize}[noitemsep]
\item Content presents universal knowledge, fundamental principles, or established facts.
\item Examples: mathematical concepts, scientific laws, historical facts, general how-to guides.
\item Information remains equally valuable and accurate over extended periods.
\end{itemize}

\textbf{6 Indeterminate}
\begin{itemize}[noitemsep]
\item Not enough context to judge timeliness.
\end{itemize}

\end{tcolorbox}

\subsection{Timeliness Mixture Experiments}

While filtering for completely evergreen content (timeliness $= 5$) yields the best standalone performance, we also evaluate a mixture-based approach that upsamples evergreen content while retaining a small proportion of time-sensitive documents. This preserves dataset diversity, avoids distributional bias, and increases effective data volume.

Table~\ref{tab:timeliness_brackets} defines the timeliness brackets used for mixture construction. In buckets 191–199, completely evergreen content (High bracket) naturally comprises 32.7\% of tokens, with the majority falling in the Medium bracket (55.6\%).

\begin{table}[h]
\centering
\caption{Timeliness bracket definitions and token proportions in buckets 191--199.}
\label{tab:timeliness_brackets}
\begin{tabular}{lcc}
\hline
Bracket & Timeliness Values & Natural Proportion \\
\hline
Low & 1, 2 & 11.7\% \\
Medium & 3, 4 & 55.6\% \\
High & 5 & 32.7\% \\
\hline
\end{tabular}
\end{table}

The mixture substantially outperforms the unfiltered baseline across all benchmarks — improving over baseline by 6.4\% on GSM8K, 7.8\% on Minerva-Math, 3.6\% on HumanEval, and 6.2\% on MBPP at $L_{10}$. However, pure $T_5$ filtering consistently outperforms the mixture on all benchmarks, with the largest gap on Minerva-Math (11.8\% vs.\ 7.8\% improvement over baseline). This suggests that even a small proportion of time-sensitive content (10\% Low in the mixture) introduces enough conflicting signal to measurably degrade mathematical reasoning training. For practitioners who require higher data volume, the mixture remains a viable option — but the pure evergreen filter is the stronger signal.

\begin{figure}[t]
\centering
\begin{subfigure}[b]{\linewidth}
\includegraphics[width=\linewidth]{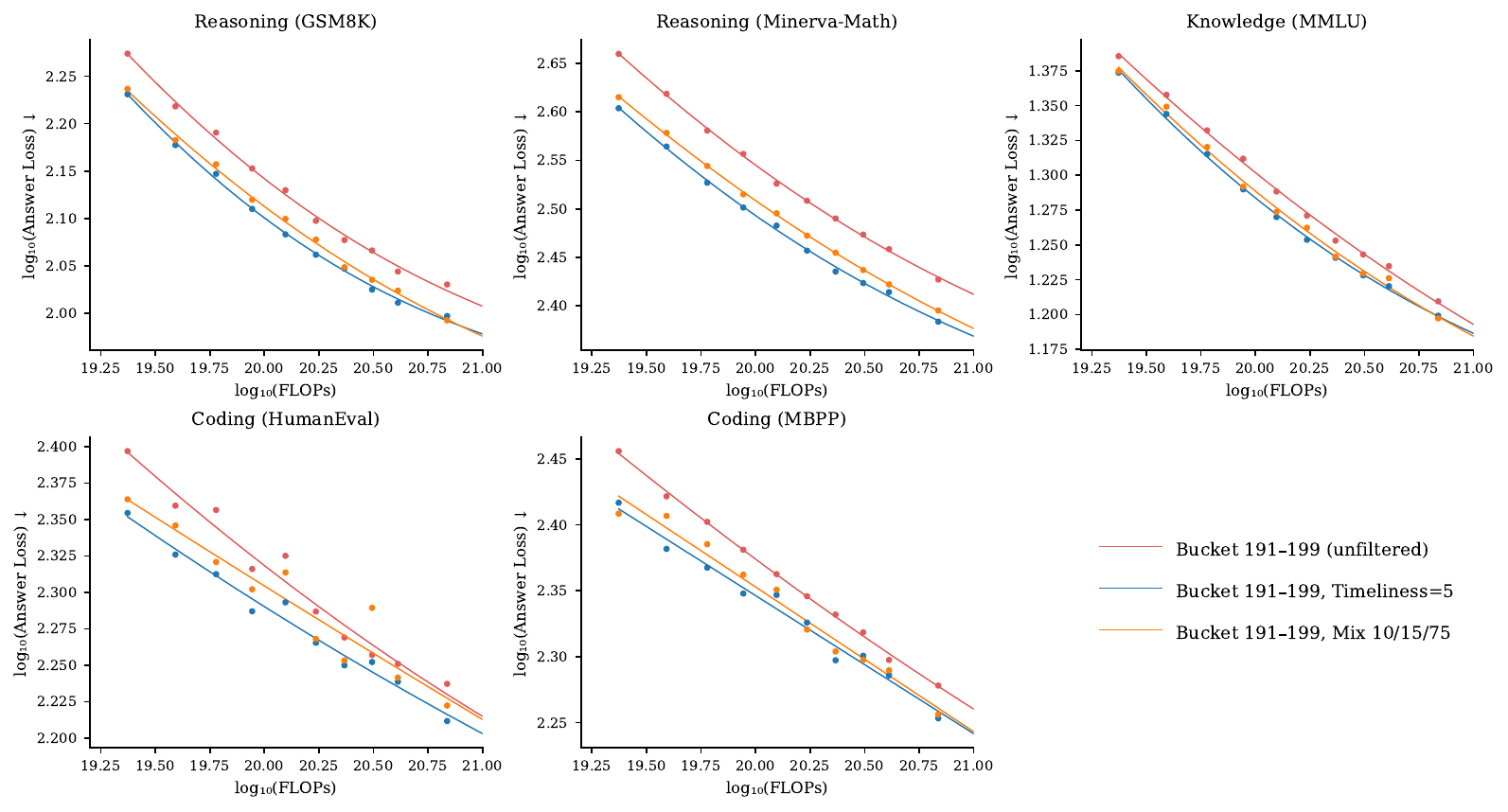}
\end{subfigure}
\caption{Scaling-law curves comparing the unfiltered bucket 191–199 baseline against pure timeliness filtering ($T{=}5$) and a mixture-based approach (10\% Low / 15\% Medium / 75\% High timeliness) on the same tier. The mixture substantially outperforms the unfiltered baseline across all benchmarks, confirming that shifting the distribution toward evergreen content is beneficial even without hard filtering. However, pure $T{=}5$ filtering consistently outperforms the mixture — with the largest gap on Minerva-Math (11.8\% vs. 7.8\% improvement over baseline) — suggesting that even a small proportion of time-sensitive content introduces enough conflicting signal to measurably degrade reasoning training.}
\label{fig:timeliness_mixture}
\end{figure}

\section{Cultural Specificity Dimension}

\subsection{LLM Annotation Prompt}
\label{app_cultural_prompt}
\begin{tcolorbox}[title=Cultural Specificity Annotation Prompt]
\small
\textbf{\#\# [Custom] Cultural Context Taxonomy}

\textbf{\#\#\# Taxonomy Definition}

\textbf{1 Highly Localized}
\begin{itemize}[noitemsep]
\item Content is deeply tied to a specific region, country, or culture.
\item Contains clear references to local events, customs, places, or norms.
\item Makes sense primarily within that specific context.
\item Examples: regional news stories, local restaurant reviews, cultural festival descriptions, local government notices.
\end{itemize}

\textbf{2 Mostly Local/Regional}
\begin{itemize}[noitemsep]
\item Content is clearly for a particular region or culture.
\item Parts may be applicable to a broader audience, but primary relevance is local.
\item Examples: national laws and regulations, country-specific educational policies, regional business practices.
\end{itemize}

\textbf{3 Balanced Local-Global}
\begin{itemize}[noitemsep]
\item Content blends local and global perspectives.
\item Includes specific regional references while conveying globally relevant ideas.
\item Examples: global issues explained through local examples, international comparisons, culturally-specific approaches to universal topics.
\end{itemize}

\textbf{4 Global but Contextualized}
\begin{itemize}[noitemsep]
\item Content is globally relevant but reflects implicit cultural assumptions or perspectives.
\item May reference specific countries/regions as examples but discusses universal topics.
\item Examples: business advice using Western norms, global topics with region-specific examples.
\end{itemize}

\textbf{5 Completely Universal}
\begin{itemize}[noitemsep]
\item Content is entirely culture-neutral with no regional specificity.
\item Applicable across cultures and regions.
\item Examples: mathematical concepts, formal scientific principles, universal logical reasoning, abstract theoretical frameworks.
\end{itemize}

\textbf{6 Indeterminate}
\begin{itemize}[noitemsep]
\item Not enough context to judge cultural specificity.
\end{itemize}

\end{tcolorbox}

\subsection{Scaling-law curves}

\begin{figure}[t]
\centering
\begin{subfigure}[b]{\linewidth}
\includegraphics[width=\linewidth]{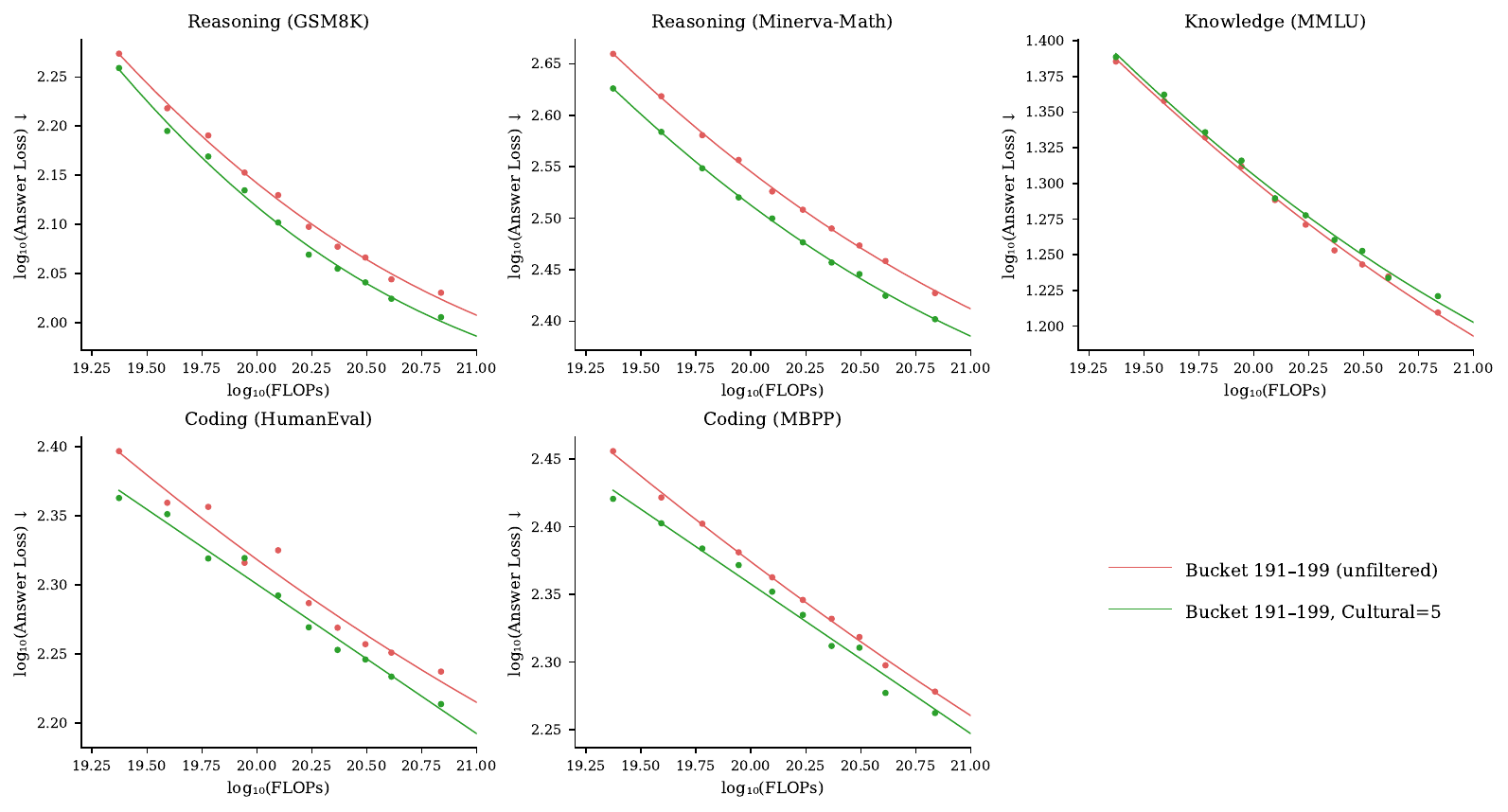}
\end{subfigure}
\caption{Scaling-law curves comparing unfiltered buckets 191–199 against subsets filtered to cultural specificity = 5 (completely universal). Filtering for completely universal content consistently improves performance across reasoning, math, and coding benchmarks. The one exception is MMLU, where the unfiltered baseline shows a marginal advantage at higher compute, attributable to knowledge breadth reduction: MMLU tests diverse domains including culturally-contextualized topics such as law, social sciences, and regional history. Cultural specificity thus presents a more nuanced tradeoff than timeliness: it is a reliable signal for reasoning and coding capability but comes at a small cost to broad factual knowledge coverage.}
\label{fig:cultural}
\end{figure}

Figure \ref{fig:cultural} compares the unfiltered buckets 191–199 against subsets filtered to cultural specificity.

\section{Cross-Analysis with Document Types and Content Dimensions}
\label{appendix_sec_cross_analysis}

This section provides detailed cross-analysis tables between the new taxonomy dimensions and existing dimensions. All distributions are computed on quality buckets 191–199 and reported as percentage of tokens.

\subsection{Timeliness}
Table~\ref{tab:timeliness_doctype} shows the distribution of the four most informative coarse document types (17-category taxonomy) across timeliness values.
Table~\ref{tab:timeliness_doctype_fine} shows the distribution across selected fine-grained document types (25-category taxonomy).
Table~\ref{tab:timeliness_reasoning} shows the reasoning depth distribution across timeliness values.

\paragraph{Cross-analysis with document types.}
The distribution of document types across timeliness values validates that the dimension captures meaningful structural differences in content. Highly time-sensitive content is dominated by News/Editorial documents (88.9\%), while completely evergreen content is dominated by Reference/Educational documents (84.3\%), with Academic/Research content comprising a further 5.9\% and News/Editorial reduced to just 0.6\%. 
Using the more granular 25-category document type taxonomy, completely evergreen content reveals a coherent cluster of durable, high-value document types: Knowledge Articles (25.6\%), Tutorials (19.6\%), Q\&A Forums (16.6\%), Academic Writing (10.0\%), and Documentation (6.5\%) together account for 78.3\% of tokens — compared to News Articles (85.4\%) and Organizational News (5.0\%) dominating the highly time-sensitive end. 
The transition is particularly clear for Tutorials and Q\&A Forums, which show monotonically increasing proportions from timeliness 1 to 5 
confirming that the timeliness classifier correctly identifies these inherently durable document types as evergreen.
These patterns validate that the six-level scale captures meaningful distinctions across content types.

\textbf{Cross-analysis with reasoning depth. }
The relationship between timeliness and reasoning depth provides a mechanistic explanation for why timeliness filtering improves downstream performance on reasoning benchmarks. Highly time-sensitive content (value 1) is dominated by Basic reasoning (58.0\%) and No Reasoning (35.6\%), with Intermediate and Advanced reasoning comprising only 6.4\% combined. Completely evergreen content (value 5) shows a markedly different profile: Intermediate reasoning (26.8\%) and Advanced reasoning (5.6\%) together account for 32.4\% of tokens. Filtering for evergreen content thus favors documents that require multi-step reasoning, while discarding the news and event coverage that dominates the low-reasoning end of the spectrum.

\subsection{Cultural Specificity}
Table~\ref{tab:cultural_doctype} shows the document type distribution across cultural specificity values. 
Completely universal content (value 5) concentrates strongly in Reference/Educational (72.1\%) and Academic/Research (17.2\%) documents, with News/Editorial reduced to 1.8\%. Highly localized content (value 1) shows a more diverse distribution — Reference/Educational (45.9\%), News/Editorial (26.5\%), and Academic/Research (11.9\%) — reflecting that locally-specific content spans multiple genres including local news, regional academic work, and culturally-specific reference material. The pattern confirms that universal content is predominantly formal reference and scientific material, consistent with the dimension's intended semantics.

\begin{table}[t]
\centering
\small
\caption{Document type distribution (\% of tokens) across timeliness values for buckets 191-199 data. 
The composition inverts monotonically from News/Editorial-dominated 
time-sensitive content to Reference/Educational-dominated evergreen content.}
\label{tab:timeliness_doctype}
\begin{tabular}{lcccc}
\toprule
\textbf{Timeliness} & \textbf{News/Editorial} & \textbf{Academic/Research} & \textbf{Ref./Educational} & \textbf{Social/Forum} \\
\midrule
Highly Time-Sensitive  & 88.9 & 0.1  & 8.6  & 0.6 \\
Pred. Time-Sensitive   & 72.7 & 2.7  & 10.9 & 2.2 \\
Balanced               & 35.9 & 5.7  & 47.5 & 2.2 \\
Pred. Evergreen        &  4.5 & 25.9 & 55.7 & 1.8 \\
Completely Evergreen   &  0.6 & 5.9  & 84.3 & 1.8 \\
\bottomrule
\end{tabular}
\end{table}

\begin{table}[t]
\centering
\small
\caption{Fine-grained document type distribution (\% of tokens) across timeliness values
(25-category taxonomy, selected types) for buckets 191-199 data. The composition transitions monotonically
from news-dominated time-sensitive content to a coherent cluster of durable,
high-value document types at the evergreen end.}
\label{tab:timeliness_doctype_fine}
\setlength{\tabcolsep}{4pt}
\begin{tabular}{lcccccccc}
\toprule
\textbf{Timeliness} & \textbf{News} & \textbf{News} & \textbf{Knowledge} & \textbf{Tutorial} & \textbf{Q\&A} & \textbf{Academic} & \textbf{Docs.} & \textbf{Personal} \\
                    & \textbf{Article} & \textbf{(Org.)} & \textbf{Article} & & \textbf{Forum} & \textbf{Writing} & & \textbf{Blog} \\
\midrule
Highly Time-Sensitive & 85.4 &  5.0 &  1.3 &  0.0 &  0.0 &  0.0 & 0.9 & 0.2 \\
Pred. Time-Sensitive  & 53.7 & 20.5 &  3.7 &  0.7 &  0.5 &  1.9 & 1.1 & 2.0 \\
Balanced              & 22.5 & 11.5 & 21.7 &  2.6 &  1.2 &  4.1 & 0.7 & 7.5 \\
Pred. Evergreen       &  3.0 &  1.5 & 24.6 &  4.8 &  6.4 & 25.2 & 5.2 & 2.5 \\
Completely Evergreen  &  0.4 &  0.0 & 25.6 & 19.6 & 16.6 & 10.0 & 6.5 & 1.7 \\
\bottomrule
\end{tabular}
\end{table}

\begin{table}[t]
\centering
\small
\caption{Reasoning depth distribution (\% of tokens) across timeliness values.
Completely evergreen content contains five times the proportion of 
Intermediate+Advanced reasoning compared to highly time-sensitive content 
(32.4\% vs.\ 6.3\%).}
\label{tab:timeliness_reasoning}
\begin{tabular}{lcccc}
\toprule
\textbf{Timeliness} & \textbf{No Reasoning} & \textbf{Basic} & \textbf{Intermediate} & \textbf{Advanced} \\
\midrule
Highly Time-Sensitive  & 35.6 & 58.0 &  6.3 & 0.0 \\
Pred. Time-Sensitive   & 16.9 & 57.9 & 23.8 & 1.4 \\
Balanced               & 12.9 & 55.2 & 29.6 & 2.3 \\
Pred. Evergreen        &  9.8 & 49.8 & 23.0 & 17.1 \\
Completely Evergreen   & 13.2 & 54.3 & 26.8 & 5.6 \\
\bottomrule
\end{tabular}
\end{table}

\begin{table}[t]
\centering
\small
\caption{Educational level distribution (\% of tokens) across timeliness values.
Time-sensitive content is overwhelmingly General Audience (93.4\%);
evergreen content distributes more evenly across educational levels,
spanning basic how-to guides through advanced scientific content.}
\label{tab:timeliness_edu}
\begin{tabular}{lcccc}
\toprule
\textbf{Timeliness} & \textbf{General Audience} & \textbf{High School} & \textbf{Undergraduate} & \textbf{Graduate/Expert} \\
\midrule
Highly Time-Sensitive  & 93.4 &  5.2 &  0.6 & 0.8 \\
Pred. Time-Sensitive   & 70.9 & 17.5 &  7.1 & 4.5 \\
Balanced               & 63.7 & 20.6 & 11.8 & 3.9 \\
Pred. Evergreen        & 39.7 & 18.4 & 14.6 & 27.1 \\
Completely Evergreen   & 34.4 & 33.9 & 25.8 & 5.8 \\
\bottomrule
\end{tabular}
\end{table}

\begin{table}[t]
\centering
\small
\caption{Document type distribution (\% of tokens) across cultural specificity values.
Universal content concentrates in Reference/Educational and Academic/Research;
localized content spans a broader mix of genres.}
\label{tab:cultural_doctype}
\begin{tabular}{lcccc}
\toprule
\textbf{Cultural Specificity} & \textbf{News/Editorial} & \textbf{Academic/Research} & \textbf{Ref./Educational} & \textbf{Social/Forum} \\
\midrule
Highly Localized           & 26.5 & 11.9 & 45.9 & 1.4 \\
Mostly Local/Regional      & 25.6 & 11.5 & 42.7 & 0.6 \\
Balanced Local-Global      & 12.3 & 24.5 & 51.5 & 1.4 \\
Global but Contextualized  & 35.7 &  8.6 & 37.7 & 4.1 \\
Completely Universal       &  1.8 & 17.1 & 72.1 & 1.9 \\
\bottomrule
\end{tabular}
\end{table}

\section{Pairwise NMI Between Taxonomy Dimensions}

Table~\ref{tab:nmi_matrix} reports pairwise normalized mutual information (NMI) between all taxonomy dimensions, computed on quality buckets 191--199. NMI is computed using the formula $\text{NMI}(X,Y) = \frac{I(X;Y)}{\sqrt{H(X) \cdot H(Y)}}$, where $I(X;Y)$ is mutual information and $H(\cdot)$ denotes entropy.

The two newly introduced dimensions — timeliness and cultural specificity — show low pairwise NMI with all existing dimensions (timeliness: max NMI $= 0.21$ with document type; cultural specificity: max NMI $= 0.11$ with timeliness), confirming that they capture genuinely novel signal not already encoded in the taxonomy. The highest pairwise NMI overall is between Bloom cognitive and Bloom knowledge ($0.41$), which is expected given their shared theoretical framework.

\begin{table}[h]
\centering
\caption{Pairwise NMI between taxonomy dimensions on quality buckets 191--199. Lower values indicate more orthogonal dimensions.}
\label{tab:nmi_matrix}
\resizebox{\textwidth}{!}{%
\begin{tabular}{lcccccccccc}
\hline
 & Bl.Cog & Bl.Know & Doc.Type & Extr.Art & Miss.Con & Reasoning & Technical & Educational & Timeliness & Cultural \\
\hline
Bloom Cognitive    & 1.000 & 0.411 & 0.175 & 0.019 & 0.008 & 0.231 & 0.039 & 0.135 & 0.041 & 0.044 \\
Bloom Knowledge    & 0.411 & 1.000 & 0.076 & 0.016 & 0.002 & 0.189 & 0.029 & 0.101 & 0.044 & 0.043 \\
Document Type      & 0.175 & 0.076 & 1.000 & 0.035 & 0.015 & 0.133 & 0.092 & 0.211 & 0.205 & 0.077 \\
Extr. Artifacts    & 0.019 & 0.016 & 0.035 & 1.000 & 0.191 & 0.014 & 0.027 & 0.027 & 0.021 & 0.013 \\
Missing Content    & 0.008 & 0.002 & 0.015 & 0.191 & 1.000 & 0.012 & 0.031 & 0.009 & 0.011 & 0.008 \\
Reasoning          & 0.231 & 0.189 & 0.133 & 0.014 & 0.012 & 1.000 & 0.114 & 0.238 & 0.029 & 0.017 \\
Technical          & 0.039 & 0.029 & 0.092 & 0.027 & 0.031 & 0.114 & 1.000 & 0.100 & 0.033 & 0.046 \\
Educational        & 0.135 & 0.101 & 0.211 & 0.027 & 0.009 & 0.238 & 0.100 & 1.000 & 0.046 & 0.028 \\
\hline
\textbf{Timeliness}     & \textbf{0.041} & \textbf{0.044} & \textbf{0.205} & \textbf{0.021} & \textbf{0.011} & \textbf{0.029} & \textbf{0.033} & \textbf{0.046} & \textbf{1.000} & \textbf{0.113} \\
\textbf{Cultural Spec.} & \textbf{0.044} & \textbf{0.043} & \textbf{0.077} & \textbf{0.013} & \textbf{0.008} & \textbf{0.017} & \textbf{0.046} & \textbf{0.028} & \textbf{0.113} & \textbf{1.000} \\
\hline
\end{tabular}%
}
\end{table}

\section{Embedding-Based Multi-Task Classifier}
\label{app:genvdisc}

\textbf{Architecture.} The model uses a shared four-layer trunk (3072$\to$3072$\to$2048$\to$1024) with BatchNorm, ReLU, and progressive dropout, branching into task-specific heads. Standard heads use a two-layer MLP (1024$\to$512$\to$output); larger heads for dimensions including FDC topic, timeliness, and cultural specificity use a three-layer MLP (1024$\to$1024$\to$512$\to$output). For hierarchical taxonomies, child-level heads receive the parent level's logit vector concatenated with trunk features, enforcing parent--child consistency.

\textbf{Training.} We optimize with AdamW (lr=$10^{-4}$, weight decay=$10^{-5}$) using cosine annealing with 1000-step warmup. Focal loss replaces standard cross-entropy to mitigate class imbalance. Task-specific loss weights emphasize quality-critical dimensions (reasoning, technical correctness: $1.5\times$) and down-weight noisier signals (extraction artifacts, missing content: $0.5\times$).

We compare the 0.5B generative classifier against a 73M-parameter discriminative MLP built on E5-multilingual embeddings~\citep{e5multilingual} (512-token context, 1024-dim). The 512-token context window is the primary explanatory variable for the 7.5\% aggregate gap: E5-multilingual truncates documents to approximately 5\% of their content before embedding. Dimensions requiring full-document reasoning (DCLM quality, document type) show gaps of 10--25 percentage points, while dimensions detectable from opening paragraphs (timeliness, cultural specificity) show gaps of only 2--3 points. Secondary labels---identifying the \emph{second}-most-relevant classification---fail catastrophically (26.0\% vs.\ 89.3\%) because they demand nuanced full-document reading. The accuracy cost is thus attributable to the context window rather than the generative-vs.-discriminative paradigm itself.

\section{Three-Tier Domain Filtering: Detailed Results}
\label{app:threetier}

Table~\ref{tab:threetierdetail} breaks down the LLM-as-Judge evaluation by quality dimension.

\begin{table}[h]
\caption{Three-tier domain filtering: per-dimension LLM-as-Judge scores (1--5 scale). 16,000 total evaluations.}
\label{tab:threetierdetail}
\centering
\small
\begin{tabular}{@{}l l cccc@{}}
\toprule
Dimension & Domain & Tier~1 & Tier~2 & Tier~3 & Instruction \\
\midrule
\multirow{4}{*}{Topic Relevance}
 & Chemistry & 4.02 & 4.58 & \textbf{4.83} & 3.16 \\
 & Comp.\ Sci. & 3.51 & 3.86 & \textbf{4.48} & 3.67 \\
 & Econ. & 2.58 & 2.61 & \textbf{3.38} & 2.73 \\
 & Physics & 4.28 & 4.56 & \textbf{4.86} & 3.17 \\
\midrule
\multirow{4}{*}{Technical Depth}
 & Chemistry & 2.83 & 3.53 & \textbf{4.05} & 2.66 \\
 & Comp.\ Sci. & 2.48 & 2.86 & \textbf{3.53} & 2.80 \\
 & Econ. & 2.29 & 2.41 & \textbf{3.04} & 2.39 \\
 & Physics & 2.92 & 3.49 & \textbf{4.04} & 2.63 \\
\midrule
\multirow{4}{*}{Educational Value}
 & Chemistry & 2.94 & 3.65 & \textbf{4.13} & 2.71 \\
 & Comp.\ Sci. & 2.58 & 2.96 & \textbf{3.56} & 2.95 \\
 & Econ. & 2.55 & 2.60 & \textbf{3.37} & 2.51 \\
 & Physics & 3.13 & 3.64 & \textbf{4.10} & 2.66 \\
\midrule
\multirow{4}{*}{Content Quality}
 & Chemistry & 3.43 & 3.72 & \textbf{4.16} & 3.26 \\
 & Comp.\ Sci. & 3.25 & 3.38 & \textbf{3.63} & 3.28 \\
 & Econ. & 3.42 & 3.47 & \textbf{3.87} & 3.14 \\
 & Physics & 3.47 & 3.70 & \textbf{4.12} & 3.12 \\
\midrule
\multirow{4}{*}{\parbox{2.5cm}{Reasoning\\Complexity}}
 & Chemistry & 2.31 & 3.11 & \textbf{3.83} & 2.45 \\
 & Comp.\ Sci. & 2.09 & 2.48 & \textbf{3.30} & 2.51 \\
 & Econ. & 2.44 & 2.51 & \textbf{3.36} & 2.28 \\
 & Physics & 2.67 & 3.27 & \textbf{3.95} & 2.50 \\
\bottomrule
\end{tabular}
\end{table}

\section{Pass 1 and Pass 2 Results}

Figure \ref{fig:pass_1} shows Pass 1 Individual dimension value results: Percentage change in answer loss relative to the unfiltered bucket 191–199 baseline for each individual taxonomy dimension value.
Figure \ref{fig:pass_2} shows Pass 2 Compound filter results: Percentage change in answer loss relative to the unfiltered bucket 191–199 baseline for compound conjunctive and disjunctive filter configurations.
Table \ref{tab:filters_retentions} shows the Pass 2 compound filter definitions and data retention rates on buckets 191--199.

\begin{table}[h]
\centering
\caption{Pass 2 compound filter definitions and data retention rates on buckets 191--199. $B_C$: Bloom Cognitive $=$ Apply; $B_K$: Bloom Knowledge $=$ Procedural; $E$: Educational level; $R$: Reasoning depth; $T$: Timeliness; $C$: Cultural specificity.}
\label{tab:filters_retentions}
\small
\begin{tabular}{llr}
\toprule
\textbf{Filter} & \textbf{Definition} & \textbf{Retention (\%)} \\
\midrule
F1  & $(B_C \vee B_K) \wedge E{\geq}2$                              & 15.29 \\
F2  & $(B_C \vee B_K) \wedge E{\geq}2 \wedge T{=}5$                & 9.03  \\
F3  & $(B_C \vee B_K) \wedge T{=}5$                                 & 10.84 \\
F4  & $B_C \vee B_K \vee T{=}5$                                     & 42.54 \\
F5  & $E{\geq}2 \wedge R{\geq}3$                                    & 33.45 \\
F6  & $B_C \vee B_K \vee E{\geq}2$                                  & 62.69 \\
F7  & $(B_C \vee B_K \vee E{\geq}2) \wedge T{=}5$                  & 23.27 \\
F8  & $E{\geq}2 \wedge R{\geq}3 \wedge T{=}5$                      & 10.31 \\
F9  & $(B_C \vee B_K \vee R{\geq}3) \wedge T{=}5$                  & 16.34 \\
F10 & $(B_C \vee B_K \vee R{\geq}3) \wedge T{=}5 \wedge E{\geq}2$  & 14.27 \\
F11 & $R{\geq}3 \wedge T{=}5$                                       & 10.62 \\
F12 & $T{=}5 \wedge C{=}5$                                          & 22.56 \\
\bottomrule
\end{tabular}
\end{table}

\end{document}